\documentclass[a4paper, twoside, 12pt]{article}

\usepackage[noblocks]{authblk} %% Affiliations with superscripts (1,2,3)
\usepackage[numbers]{natbib}
\usepackage{graphicx}
\usepackage{fancyhdr}
\usepackage[colorlinks=true, linkcolor=MidnightBlue, urlcolor=MidnightBlue, citecolor=PineGreen]{hyperref}
\usepackage[tableposition=top, justification=justified, textfont=it]{caption}
\usepackage{titlesec}
\usepackage{multicol}
\usepackage{multirow}
\usepackage{afterpage}
\usepackage{lipsum}
\usepackage[dvipsnames, x11names, table]{xcolor}
\usepackage{colortbl,hhline}
\usepackage[inner=4cm, outer=4cm, top=2.8cm, bottom=2.8cm, marginparsep=0cm, marginparwidth=0cm, includehead, includefoot]{geometry}
\usepackage{setspace}
\usepackage{array}
\usepackage{bm}
\usepackage{tikz}
\usepackage{float}
\usepackage[framemethod]{mdframed}
\usepackage{amsmath}
\usepackage{amssymb}
\usepackage{booktabs}
\usepackage[textsize=tiny]{todonotes}

\titleformat{\section}{\Large \bfseries \centering \scshape}{\thesection.}{0.3em}{}[{\titlerule[0.5pt]}]

\definecolor{shadecolor}{RGB}{230,230,230}
\newcommand{\mybox}[1]{\par\noindent\colorbox{shadecolor}
{\parbox{\dimexpr\textwidth-2\fboxsep\relax}{#1}}}

\titleformat{\subsection}{\large \bfseries \mybox}{\thesubsection}{1em}{}

\titleformat{\subsubsection}{\itshape}{\thesubsubsection.}{0.3em}{}

\renewenvironment{abstract}
{\vskip 2.5ex {\large\bf\noindent Abstract}\vspace{0.7ex} \\ %
  \bgroup\noindent\ignorespaces}%
{\par\egroup\vskip 2.5ex}

\newenvironment{keywords}
{\bgroup\leftskip 20pt\rightskip 20pt \small\noindent{\bf Keywords:} }%
{\par\egroup\vskip 10ex}

\fancypagestyle{firstpage}{%
  
  \fancyhead[]{}
  \fancyfoot[C]{}
  \fancyfoot[L]{\footnotesize To appear in \\
  O. Colliot (Ed.), \textit{Machine Learning for Brain Disorders}, Springer\\
  }

}

\fancypagestyle{otherpages}{%
  
  \fancyhead[LE]{\thepage}
  \fancyhead[RE]{\runningauthor}
  \fancyhead[LO]{\runningheadtitle}
  \fancyhead[RO]{\thepage}
  \fancyfoot[L]{\footnotesize  \textit{Machine Learning for Brain Disorders} 
  }
  \fancyfoot[C]{}
}

\makeatletter
\renewcommand{\maketitle}{\bgroup\setlength{\parindent}{0pt}

\begin{flushright}
  \color{MidnightBlue}
\end{flushright}

\vspace{-2cm}

\begin{flushleft}
    \setstretch{2.0} %% allows a nicer formatting of the title by adding more space between lines
    \textbf{\color{MidnightBlue}\huge\@title}
\end{flushleft}

\vspace{0.15in}

\begin{flushleft}
    \textbf{\bfseries \large\@author}
\end{flushleft}\egroup
}

\makeatother
\DeclareCaptionLabelFormat{labelformat}{\color{MidnightBlue}\bfseries #1 #2}
\renewcommand{\bibpreamble}{\scriptsize \begin{multicols}{2}}
\renewcommand{\bibpostamble}{\end{multicols}}

\mdfsetup{skipabove=\topskip, skipbelow=\topskip}

\newcounter{nicebox}
\newenvironment{nicebox}[1][]{%
    \refstepcounter{nicebox}%
    \ifstrempty{#1}%
    {\mdfsetup{%
        frametitle={%
            \tikz[baseline=(current bounding box.east),outer sep=0pt]
            \node[anchor=east,rectangle,fill=blue!20]
            {\strut Theorem~\thetheo};}}
    }%
    {\mdfsetup{%
        frametitle={%
            \tikz[baseline=(current bounding box.east),outer sep=0pt]
            \node[anchor=east,rectangle,fill=blue!20]
            {\strut Box~\thenicebox:~#1};}}%
    }%
    \mdfsetup{innertopmargin=10pt,linecolor=blue!20, linewidth=2pt,topline=true, frametitleaboveskip=\dimexpr-\ht\strutbox\relax,}
    \begin{mdframed}[]\relax%
    }{\end{mdframed}}

\newfloat{floatbox}{thp}{lop}
\floatname{floatbox}{Box}

\DeclareMathAlphabet{\mathsfit}{\encodingdefault}{\sfdefault}{m}{sl}
\SetMathAlphabet{\mathsfit}{bold}{\encodingdefault}{\sfdefault}{bx}{n}

\usepackage{smartdiagram}
\usepackage{enumitem}

\definecolor{antiquewhite}{rgb}{0.98, 0.92, 0.84}
\definecolor{beaublue}{rgb}{0.74, 0.83, 0.9}

\begin{document}

%%%%%%%%%%%%%%%%%%%%%%%%%%%%%%%%%%%%%%%%%%%%%%%%%%%%%%%%%%%%%%%%%
%           I N F O R M A T I O N   T O   C H A N G E
%%%%%%%%%%%%%%%%%%%%%%%%%%%%%%%%%%%%%%%%%%%%%%%%%%%%%%%%%%%%%%%%%

% Author name displayed in the running head
\newcommand{\runningauthor}{Colliot et al} 

% Title displayed in the running head
\newcommand{\runningheadtitle}{Reproducibility}

% Chapter number
\newcommand{\chapternumber}{20}
\renewcommand{\chapternumber}{} % XXX: to remove after preprint

% E-mail address of the corresponding author
\newcommand{\emailaddress}{olivier.colliot@cnrs.fr}

% Title of the chapter
\title{Reproducibility in machine learning for medical imaging} 

% Authors' names and affiliation numbers
\author[1,*]{Olivier Colliot}  % Use the symbol '*' for the corresponding author
\author[2]{Elina Thibeau-Sutre}  
\author[1]{Ninon Burgos}  

% Affiliations
\affil[1]{Sorbonne Universit\'e, Institut du Cerveau - Paris Brain Institute - ICM, CNRS, Inria, Inserm, AP-HP, H\^opital de la Piti\'e-Salp\^etri\`ere, F-75013, Paris, France}
\affil[2]{Department of Applied Mathematics, Technical Medical Centre, University of Twente, Enschede, The Netherlands}
%%%%%%%%%%%%%%%%%%%%%%%%%%%%%%%%%%%%%%%%%%%%%%%%%%%%%%%%%%%%%%%%%
\affil[*]{Corresponding author: e-mail address: \href{mailto:\emailaddress}{\emailaddress}}

\maketitle

% Restore the geometry and change the page style for the other pages
\afterpage{\aftergroup\restoregeometry}
\pagestyle{otherpages}

% Abstract
\begin{abstract}
Reproducibility is a cornerstone of science, as the replication of findings is the process through which they become knowledge.  It is widely considered that many fields of science are undergoing a reproducibility crisis. This has led to the publications of various guidelines in order to improve research reproducibility.

This didactic chapter intends at being an introduction to reproducibility for researchers in the field of machine learning for medical imaging.  We first distinguish between different types of reproducibility. For each of them, we aim at defining it, at describing the requirements to achieve it and at discussing its utility.  The chapter ends with a discussion on the benefits of reproducibility and with a plea for a non-dogmatic approach to this concept and its implementation in research practice.
\end{abstract}

% Keywords

\definecolor{mypurple}{rgb}{.4,.0,.6}
\newcommand{\preprint}[1]{\textcolor{mypurple}{\textbf{\sffamily #1}}}
\fbox{\parbox{0.90\linewidth}{\preprint{Disclaimer: this is a working
paper, and is still work in progress. For any comments or missing
references, please email  at 
olivier.colliot@cnrs.fr}}}

%TO REMOVE FOR FINAL VERSION
\newpage

\section{Introduction}
\label{sec:intro}

Reproducibility is at the core of the scientific method. In its general and most common meaning, it corresponds to the ability to reproduce the findings of a given experimental study. This is a necessary (but not sufficient) condition for a scientific statement to become accepted as new knowledge. Let's illustrate this with a simple example, considering the following statement: ``the volume of the hippocampus is, on average, smaller in patients with Alzheimer's disease (AD) than in healthy people of comparable age''. Such statement was the conclusion of studies which measured such volume from magnetic resonance images (MRI). To the best of our knowledge, the first study to assert this was that of Seab et al~\cite{seab1988quantitative}. This was later reproduced by many other studies (e.g.  \cite{lehericy1994amygdalohippocampal, jack1997medial}). It is now widely accepted, which would not have been the case if the study had proven impossible to reproduce. Note that, as stated above, this is a {\sl necessary} but not a {\sl sufficient} condition. Indeed, there could be other reasons for such statement not to be considered as knowledge. For instance, let's imagine that some other researchers discover that there is an artifact that is systematically present in the MRI of patients with AD and which leads to erroneous volume estimation. Then, the statement could not be considered new knowledge even though it had been reproduced several times.

Machine learning (ML) is, in part, an experimental science. This is not the case of the entirety of the discipline, part of which is theoretical (for instance mathematical proofs of convergence or of approximation capabilities of different classes of models) or methodological (the invention of a new approach). Nevertheless, since ML ultimately aims at solving practical problems, its experimental component is essential. Typically, one would want to be able to make statements of the type described above from an experimental study. Here is an example of such statement: ``this ML model (for instance a specific convolutional neural network [CNN] architecture), using MRI data as input, is capable of classifying AD patients and healthy controls with an accuracy superior to 80\%''. In order to end an article with such a statement, one needs to conduct an experimental study. For such findings to become knowledge, it needs to be subsequently reproduced. Of course, this statement is unlikely to be universal and one would want to know under which conditions it holds: for instance, is it restricted to a specific class of MRI scanners? to specific disease stages? to specific age ranges?

\begin{floatbox}[htp]
    \begin{nicebox}[Glossary]
        \label{box:glossary}
        
        \footnotesize
The readers will find the definition of the terms we used in the present document.  
	  \begin{itemize}
	  \itemsep-0.3em 
	      \item {\bf Reproducibility,  replicability,  repeatability.} In the present document,  these will be used as synonyms of reproducibility. 
	      \item {\bf Original study.} Study that first showed a finding.
		  \item {\bf Replication study.} Study that subsequently aimed at replicating an original study,  with the hope to support its findings. 
		 \item {\bf Research artifact.}  Any output of scientific research: papers,  code,  data,  protocols\ldots. Not to be confused with imaging artifacts which are defects of imaging data.
		\item {\bf Claims.} The conclusions of a study.  Basically  a set of statements describing the results and a set of limitations which delineate the boundaries within which the claims are stated (the term ``claim'' is here used in the broad scientific sense not with the specific meaning it has in the context of regulation of medical devices although the two may be related).
		\item{\bf Limitations.} A set of restrictions under which the claims may not hold (usually because the corresponding settings have not been explored).
		%\item {\bf Reproducibility,  replicability,  repeatability.} In the present document,  these will be used as synonyms of reproducibility.   
		\item {\bf Method.}  The ML approach described in the paper,  independently of its implementation.
		\item   {\bf Code.}  The implementation of the method.   
		\item {\bf Software dependencies.} Other software packages that the main code relies on and which are necessary for its execution.  
		\item {\bf Public data.}  Data that can be accessed by anybody  with no or little restriction (for instance the data hosted at \url{https://openneuro.org}). 
		\item {\bf Semi-public data.} Data which requires approval of a research project (for instance the Alzheimer's Disease Neuroimaging Initiative [ADNI] \url{http://www.adni-info.org}). The researchers can then use the data only for the intended research purpose and cannot redistribute it.
%	\item {\bf Trained models}.  ML models  trained  in the original study. 
				\item {\bf Data split.} Separation into training, validation and test sets.  
				
								\item {\bf Data leakage.} Faulty procedure which has led information from the training set to leak into the test set.  See \cite{varoquaux2022evaluating,thibeau2022clinicadl} for details. 
									\item {\bf Error margins.} A general term for providing the precision of the performance estimates (e.g.  standard-error or confidence intervals).
										\item {\bf Researcher degrees of freedom.} Number of different components (e.g. different architectures, hyperparameter values, subsamples\ldots) which have been tried  before arriving to the final method~\cite{simmons2011false}.   Too many degrees of freedom tend to produce methods that do not generalize.
								\item {\bf p-hacking.} A bad practice that involves too many degrees of freedom and which consists is trying many different statistical procedures until a significant p-value is found.
					\item {\bf Acquisition settings.} Factors that influence the scan of a given patient (imaging device,  acquisition paratemeters, image quality).
					\item {\bf Image artifacts.} Defects of a medical image,  these may include noise,  field heterogeneity,  motion artifacts and others.
					\item {\bf Preregistration.} The deposit of the study protocol prior to performing the study. Limits degrees of freedom and increases likelihood of robust findings.

	  \end{itemize}  
   \end{nicebox}
\end{floatbox}

In the examples above,  we have actually illustrated only one on the many possible meanings of reproducibility: the addition of new evidence to support a scientific finding of an {\sl original} study  through reproduction under different experimental conditions (see \autoref{box:glossary} for a glossary of some of the key concepts used).  However, it is also used for very different meanings. In computational sciences, it is often used for the ability to exactly reproduce the results (i.e. the exact numbers) in a given study.  In sciences which aim at providing measurements (as is often the case in medical imaging),  the word may be used to describe the variability of a given measurement tool under different acquisition settings. We shall provide more details on these different meanings in Section~\ref{sec:polysemy}. Finally,  the topics of reproducibility and open science are obviously related since the latter favors the former. However,  open science encompasses a broader objective which is to make all {\sl research artifacts} (code, data, papers\ldots) openly available for the benefit of the whole society. Conversely,  open research may still be unreproducible (e.g.  because it has relied on faulty statistical procedures).

There has been increasing concern that science is undergoing a reproducibility crisis~\cite{baker20161, gundersen2020reproducibility,ioannidis2005most,begley2015reproducibility}. This is present in various fields from psychology \cite{open2015estimating} to preclinical oncology research \cite{begley2013unappreciated}.  ML~\cite{sonnenburg2007need,gundersen2018state,hutson2018artificial,haibe2020transparency}, digital medicine~\cite{stupple2019reproducibility} and ML for healthcare~\cite{mcdermott2019reproducibility,beam2020challenges} are no exception. The concerns are multifaceted.  In particular, they include two substantially different aspects: the report of failures to reproduce previous studies and the observation that many papers do not provide sufficient information for reproducing their results.  It is important to have in mind that, while the two may be related, there is not a direct relationship between them: it may very well be that a paper seems to include all the necessary information for reproduction and that reproduction attempts fail (for instance,  because the original study had too many degrees of freedom and led to a method that only works on a single dataset, see Section~\ref{sec:statistical}).

Various guidelines have been proposed to improve research reproducibility. Such guidelines may be general~\cite{begley2015reproducibility} or devoted to specific fields including brain imaging~\cite{gorgolewski2016practical,nichols2017best, poldrack2017scanning} and ML for healthcare and life sciences~\cite{turkyilmaz2020reproducibility, heil2021reproducibility}.  Moreover many other papers, even though not strictly providing guidelines, provide very valuable pieces of advice for making research more trustworthy and in particular more reproducible (e.g.~\cite{gundersen2018state,mcdermott2019reproducibility,varoquaux2018cross,button2013power,
varoquaux2022machine,bouthillier2019unreproducible,langer2018collaborative}).

This chapter is  an introduction to the topic of reproducibility for researchers in the field of ML for medical imaging.  It is not meant at providing a replacement for the aforementioned previously published guidelines, that we strongly encourage the reader to refer to.

The remainder of the chapter is organized as follows. We first start by introducing different types of reproducibility (Section~\ref{sec:polysemy}). For each of them, we attempt to clearly define it, describe what are the requirements to achieve it and the benefits it can provide (Sections~\ref{sec:exact} to \ref{sec:measurement}).  All this information is given with having the field of ML for medical imaging as a target, even though part of it may apply to other fields.  Finally, we conclude with a discussion which both describes the benefits of reproducibility but also advocates for a non-dogmatic point-of-view on the topic (Section~\ref{sec:discussion}).

\section{The polysemy of reproducibility} 
\label{sec:polysemy}

The term ``reproducibility'' has been used with various meanings which may range from the exact reproduction of a study with the same material and methods, to the reproduction of a result using new experimental data, to the support of a scientific idea using a completely different experimental set-up~\cite{goodman2016does, plesser2018reproducibility}. Moreover, various terms have been introduced including reproducibility, replicability, repeatability, reliability, robustness, generalizability\ldots Some of these words, for instance reproducibility vs replicability, have even been used by some authors with opposite meanings~\cite{goodman2016does, plesser2018reproducibility}. We will not aim at assigning an unambiguous meaning to each of these words, as we find this of little interest, and will use the term ``reproducibility'',  ``replicability'' and ``repeatability'' as synonyms. On the other hand, we believe,  as many other authors~\cite{mcdermott2019reproducibility, nichols2017best, goodman2016does, mcdermott2021reproducibility}, that it is important to distinguish between different types of reproducibility. To that purpose, it is useful to have a {\sl taxonomy} of reproducibility. Below, we describe such a taxonomy. We do not claim that it is novel, as it takes inspiration from other papers~\cite{gundersen2018state, mcdermott2019reproducibility, nichols2017best, goodman2016does, mcdermott2021reproducibility} nor that it should be universally adopted. Furthermore, boundaries between different types of reproducibility are partly fuzzy. We simply hope that it will be useful for the different concepts that we subsequently introduce and that it will be well-adapted to the field of ML for medical imaging.

\begin{figure}
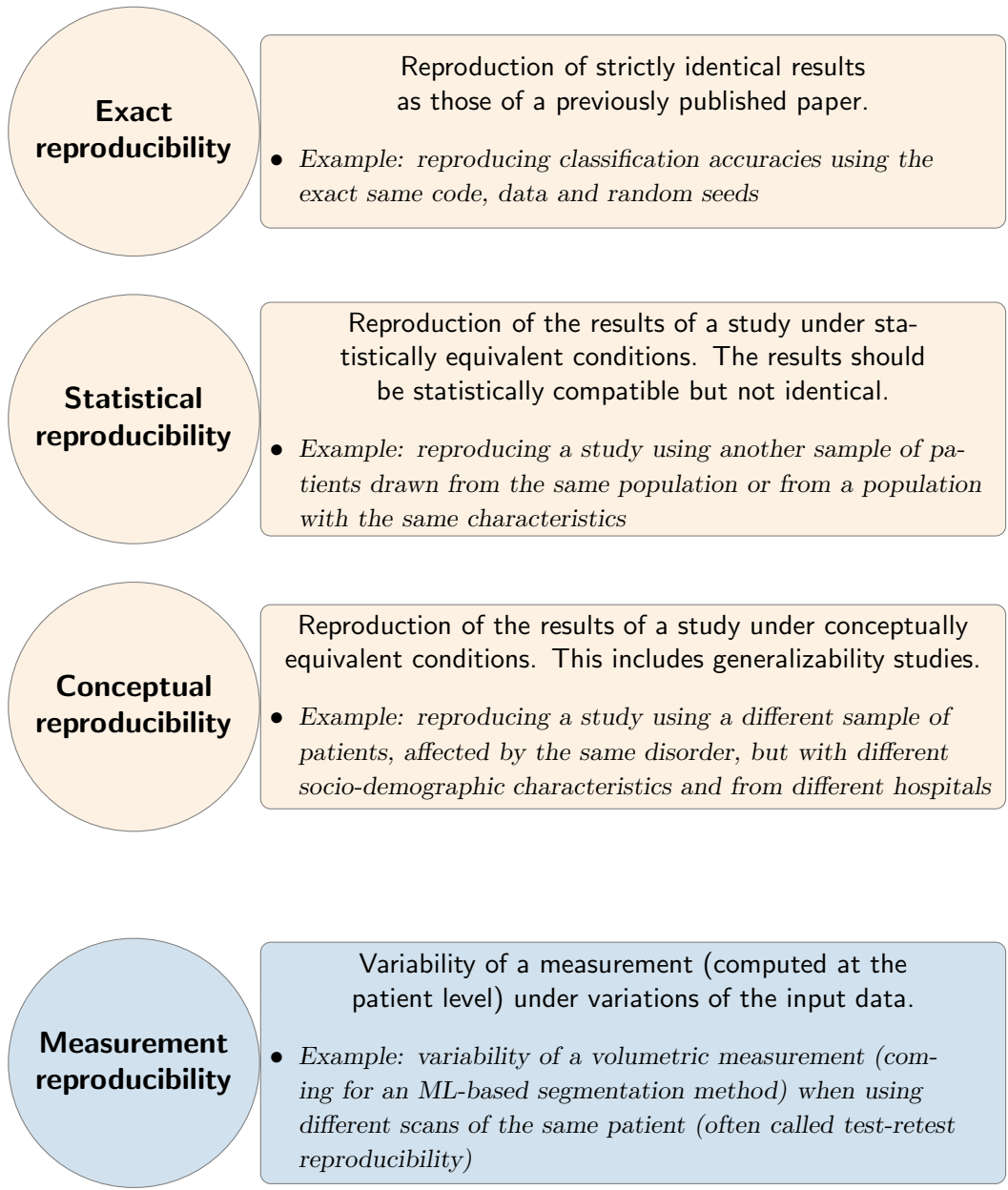

    \centering
    \sffamily
    \tikzset{no shadow/.style={
         every shadow/.style={
             fill=none,
             shadow xshift=0pt,
             shadow yshift=0pt}
     }}
     \tikzset{description/.append style={top color=\col,bottom color=\col,
     minimum height=6.5em,no shadow},
     description title/.append style={top color=\col,bottom color=\col,
     minimum height=6.5em,no shadow}}
    \smartdiagramset{
        description title font = \normalsize,
        descriptive items y sep = 4cm,
        description title width = 2.5cm,
        description title text width = 3cm,
        description width = 10cm,
        description text width = 10cm,
        set color list={antiquewhite!70,antiquewhite!70,antiquewhite!70}
    }
    \smartdiagram[descriptive diagram]{
        {\textbf{Exact\\ reproducibility},
            {Reproduction of strictly identical results as those of a previously published paper. 
            \noindent \begin{itemize}[leftmargin=1em]
                \item {\sl \footnotesize Example: reproducing classification accuracies using the exact same code,  data and random seeds}
            \end{itemize}
            }
        },
        {\textbf{Statistical \\ reproducibility}, 
            {Reproduction of the results of a study under statistically equivalent conditions. The results should be statistically compatible but not identical.
    	      \begin{itemize}[leftmargin=1em]
                \item {\sl \footnotesize Example: reproducing a study using another sample of patients drawn from the same population or from a population with the same characteristics}
    	      \end{itemize}
            }
        },
        {\textbf{Conceptual \\reproducibility}, 
            {Reproduction of the results of a study under conceptually equivalent conditions. This includes generalizability studies.
    	      \begin{itemize}[leftmargin=1em]
                \item {\sl \footnotesize Example: reproducing a study using a different sample of patients,  affected by the same disorder,  but with different socio-demographic characteristics and  from different hospitals}
    	      \end{itemize}
            }
        },
    }
    \newline
    \bigskip
    \smartdiagramset{
        description title font = \normalsize,
        descriptive items y sep = 4cm,
        description title width = 2.5cm,
        description title text width = 3cm,
        description width = 10cm,
        description text width = 10cm,
        set color list={beaublue!70}
    }
    \smartdiagram[descriptive diagram]{
        {\textbf{Measurement \\ reproducibility}, 
            {Variability of a measurement (computed at the patient level) under variations of the input data.
    	      \begin{itemize}[leftmargin=1em]
                \item {\sl \footnotesize Example: variability of a volumetric measurement (coming for an ML-based segmentation method) when using different scans of the same patient (often called test-retest reproducibility)}
            \end{itemize}
            }
        },
    }
    \caption{Different types of reproducibility. Note that, in the case of ``statistical'' and ``conceptual'' reproducibility, the terms come from~\cite{mcdermott2019reproducibility} but the exact definition provided in each corresponding section may differ.}
    \label{fig:types_reproducibility}
\end{figure}

We distinguish between four main types of reproducibility: {\bf exact reproducibility}, {\bf statistical reproducibility},  {\bf conceptual reproducibility}  and  {\bf measurement reproducibility}.  We describe those four main types in the following sections. They are also summarized in~\autoref{fig:types_reproducibility}. As will be explained below, the three first types have relationships with each other (this is why they have the same color in the figure) while the fourth is more separated.

\newpage

\section{Exact reproducibility}
\label{sec:exact}
\paragraph{What is it?}
Exact reproducibility aims at reproducing strictly identical results as those of a previously published paper.  Concretely,  this amounts to being able to reproduce tables and figures as they appear in the original paper following the same procedures as the authors.

\paragraph{What does it require?} Exact reproducibility requires to have access to all components that led to the results including,  of course,  data and code.

Access to data is obviously necessary~\cite{mcdermott2019reproducibility,gorgolewski2016practical, heil2021reproducibility}. 
Open data has been described (together with code and papers) as one of the pillars of open science~\cite{gorgolewski2016practical}.  It is widely accepted that scientific data should adhere to the FAIR (Findable, Accessible, Interoperable, Reusable) principles (please refer to \url{https://www.go-fair.org/fair-principles} and ~\cite{wilkinson2016fair} for more details). Among these principles,  accessibility is often the most difficult to adhere to for medical imaging data (or healthcare data in general). It is very common in medical papers that data is mentioned as available upon request. However, a study has showed that, when data is subsequently requested, many researchers actually do not comply with the data accessibility statement~\cite{gabelica2022many}. This is worrisome and more transparent ways of data sharing would be welcome. However, as mentioned above, such transparent sharing procedures may be difficult to put in place for healthcare data. In particular, making the data public is often difficult due to regulatory and privacy constraints \cite{mcdermott2019reproducibility}.  \citet{gorgolewski2016practical} provides useful pieces of advice to facilitate sharing but there are cases where public sharing will remain impossible.  In particular,  one must distinguish between research data (acquired as part of a research protocol), which can often be made {\sl public} or {\sl semi-public}\footnote{See glossary \autoref{box:glossary}.} provided that adequate measures have been taken at data collection (e.g. adequate participant consent), and routine clinical data (acquired as part of the routine clinical care of the patients) and which sharing can be much more complicated.  It is important that data is easily findable and that it is shared on a server which has a long-term maintenance. General purpose data repositories such as Zenodo\footnote{\url{https://zenodo.org/}} provide an excellent solution.  Another important aspect is to adhere to community standards for data organization,  so that it can easily be reused by researchers. For brain imaging, the community standard is BIDS (Brain Imaging Data Structure) \cite{gorgolewski2016brain}\footnote{\url{https://bids.neuroimaging.io/}}.  This standard is already very mature and has already been extended to incorporate other modalities such as microscopy images for instance (Microscopy-BIDS~\cite{bourget2022microscopy}). Note that there is an ongoing proposal to extend it to other organs (MIDS - Medical Imaging Data Structure~\cite{saborit2020medical}\footnote{See BIDS extension proposal (BEP) number 25 (BEP025) \url{https://bids.neuroimaging.io/get_involved.html\#extending-the-bids-specification}}). Finally, we would like to draw the attention to an important point that is often overlooked. Even when a study relies on public or semi-public data, it is absolutely necessary to specify which samples (e.g. which participants and which scans) have been used, otherwise the study is not reproducible~\cite{cuingnet2011automatic}.  Ideally, one would provide code to automatically make the data selection \cite{samper2018reproducible} in order to ease the replication.

Another key component is that the code is accessible \cite{mcdermott2019reproducibility,gorgolewski2016practical, heil2021reproducibility}.  Indeed, it would be delusional to think that exactly the same results could be obtained using a reimplementation based on information provided in the paper (even though it is good practice to provide as much detail as possible about the methods in the paper).  Theoretically,  it does not mean that the code must come with an open software license. However, doing so has many additional benefits such as allowing other researchers to use the code or parts of it for different purposes.  The code should be made according to good coding practices which include the use of a versioning system and adequate documentation~\cite{beam2020challenges}.  Furthermore,  although not strictly needed for reproducibility,  the use of continuous integration makes the code more robust and eases its long-term maintenance.  Besides, it is good to ease as much as possible the installation of dependencies \cite{heil2021reproducibility}.  This can be done with pip\footnote{\url{https://pypi.org/project/pip/}} when programming in Python.  One can also use containers such as Docker\footnote{\url{https://www.docker.com/}}.  Note that we are not saying that all these components should be present in any study or are prerequisites for good research. They constitute an ideal reproducibility goal. 

In ML,  sharing the code itself is not enough for exact reproducibility. First,  every element of the training procedure should be stored: this includes the data splits and the criteria for model selection. Moreover,  there usually are non-deterministic components so it is necessary to store random seeds~\cite{heil2021reproducibility}.  Furthermore,  software/operating system versions, the GPU model/version and threading have been deemed necessary to obtain exact reproducibility~\cite{crane2018questionable}.   The ClinicaDL software platform provides a framework for easing exact reproducibility of deep learning for neuroimaging \cite{thibeau2022clinicadl}\footnote{\url{https://clinicadl.readthedocs.io/en/latest/}}.  Although it is targeted at brain imaging, many of its components and concepts are applicable to medical imaging in general.  Also in the field of brain imaging,  NiLearn\footnote{\url{https://nilearn.github.io/stable/index.html}} facilitates the reproducibility of statistics and ML.

Finally,  it may seem obvious but, even when the code is shared,  the underlying theory of the method, all its components and implementation details need to be present in the paper~\cite{gundersen2018state}.

It is in principle possible to retrain models identically if the above elements are provided.  It nevertheless remains a good practice to share trained models,  in order to allow other researchers to check that retraining indeed led to the same results but also to save computational resources.  However,  models can be attacked to recover training data~\cite{heil2021reproducibility, carlini2021extracting}. This is not a problem when the training data is public. When it is privacy-sensitive,  methods to preserve privacy exist \cite{heil2021reproducibility, abadi2016deep}.

In medical imaging,  preprocessing and feature extraction are often critical steps  that will subsequently influence the ML results. It is thus necessary to also provide code for such parts.  Several software initiatives including BIDSApps~\cite{gorgolewski2017bids}\footnote{\url{https://bids-apps.neuroimaging.io/}} and Clinica~\cite{routier2021clinica}\footnote{\url{https://www.clinica.run/}} provide ready-to-use tools for preprocessing and feature extraction for various brain imaging modalities.  Applicable to many medical imaging modalities,  the ITK~\cite{mccormick2014itk, yoo2002engineering}~\footnote{\url{https://itk.org/}} framework provides a wide range of processing tools.  It can ease the work of researchers who do not want to spend time on preprocessing and feature extraction pipelines and focus on the ML part of their work.

\paragraph{Why is it useful?}
It has been claimed that exact reproducibility is of little interest, that pursuing it is a waste of energy of the community  and that its only possible use would be the detection of outright fraud which is rare \cite{drummond2009replicability}.  We disagree with that view.  Let's start with fraud.  It may be of low occurrence, although its exact prevalence is difficult to establish. Even so, it is of disastrous consequences as it leads to loss of trust by students,  scientists and the general public.  In particular, a  survey of 1,576 researchers indicated that $40\%$ of them believe that fraud is a factor that ``always/often'' contributes to irreproducible research and that $70\%$ of them think that it ``sometimes'' contributes~\cite{baker20161}. Exact reproducibility can certainly contribute to reduce fraud as full transparency obviously makes fraud more difficult.  Fraud remains possible (one could forge some data and share it) but it is more difficult to achieve under transparency constraints.  Fraud may be rare but errors are much more common.  The framework of exact reproducibility eases the detection of errors which is a service to science and even to the authors themselves.  In particular, it may help discover ``biases and artifacts in the data that were missed by the authors and that cannot
be discovered if the data are never made available'' \cite{heil2021reproducibility}.   Similarly,  it can lead to the discovery of wrong validation schemes, including data leakage or errors in implementation that make it inconsistent with the methodology presented in the paper.
Overall,  it may make progress slower but it will definitely make it steadier.
However, this does not mean that exact reproducibility should be aimed in all works or made a requirement for all publications (see Section~\ref{subsec:onesize}).

\section{Statistical reproducibility}
\label{sec:statistical}
\paragraph{What is it?}
Statistical reproducibility aims at reproducing findings under statistically equivalent conditions\footnote{We use the term of \citet{mcdermott2019reproducibility} although with a slightly different (more extensive) meaning. }.  The specific definition may vary but the following choices are often considered reasonable. The implementation of the method (the code) remains the same.  Random components are left random.  
Regarding the data,  the general idea is that the sample would be drawn from the same population.  One could for instance use subsamples of the original data or another subsample of a larger source population.  An interesting case is to study different data splits.  A less restrictive view of statistical reproducibility would be to use another dataset whose characteristics are similar to those of the original dataset (for instance similar age, sex, scanner distributions). Note that the boundary between statistical and conceptual reproducibility (defined in the next section) is fuzzy. We do not believe it is possible to draw exact frontiers that would delimit the statistical variations that are admissible in a statistical reproducibility study. Finally, it is important that those who conduct the statistical replication study clearly indicate which components of variability they study.

\paragraph{What does it require?} Here one needs to distinguish between two types of factors: those necessary to {\sl attempt} reproducibility and those that increase the likelihood of {\sl successful} reproducibility.

Regarding the first type,  most factors are common with those for exact reproducibility.  Code needs to be accessible so that variations coming from reimplementation do not impact the replication.  Random seeds,  GPU model or other software/execution parameters will not be set to be identical because the aim is precisely to check is the findings of the study are  statistically reproducible under such variations.  Knowing their value in the original study is nevertheless useful in order to dissect potential reasons for failed replication.  Trained models are in a similar situation: they will usually not be used for statistical replication (models will be retrained) but shall prove useful to dissect potential failures. Data accessibility is also very valuable because it will allow studying different data splits,  or subsamples.  

The above mentioned elements make it possible for other researchers to attempt statistical replication of a given study.  On the other hand,  there are features of the original study that will make such replication more likely to be successful (equivalently, one could say that the original findings are robust).  One important factor is that the original study reports error margins (reporting the standard error or equivalently a confidence interval).  It is important in this specific context because statistical reproducibility does not aim at obtaining (and cannot obtain) exactly the same results.  One wants the results to be {\sl compatible} with original ones: typically a successful replication would produce results which are within the error margin of the original study.  Beyond the topic of statistical reproducibility,  the report of error margins is of great importance in general,  in particular in the field of ML for medical imaging,  because it provides a precision on the estimates of the performance.  Unfortunately,  this practice is still too uncommon in the ML field as a whole~\cite{mcdermott2019reproducibility}.  Even worse,  it is not uncommon to find faulty interpretations of estimates.  For instance,  one should never estimate standard errors (SE) from multiple runs of a cross-validation, as the number of runs can be made arbitrarily large and as a consequence the SE arbitrarily small (see \cite{varoquaux2022evaluating}).  A very common example is papers which report empirical standard-deviation (SD) across $k$-folds.  Unlike what is quite widely believed,  this value does not allow to gauge the precision of the performance estimation.  It provides some insight on the variability of the learning procedure under variations of the training and validation sets.  Further,  keep in mind that when $k$ is small,  such gauge will be very rough.  With $k$ sufficiently large,  it is possible to assess if a ``learner'' (i.e.  an ML procedure to perform a task) is superior to another one by counting the fraction of folds on which it obtains superior performance (e.g. 75\%)~\cite{bouthillier2021accounting}.  See \cite{varoquaux2022evaluating} for more details on this question.  However,  in no case can such procedures estimate the precision of the performance of the trained model, in other words the precision of the computed biomarker or computer-aided diagnosis tool.  This requires an independent test set, from which SE and confidence intervals can be computed.

\paragraph{Why is it useful?}
Statistical replication has many merits.  First,  by reassessing ML methods using different data splits,  one can spot faulty procedures including data leakage which is prevalent in the field of medical imaging \cite{wen_convolutional_2020, samala_hazards_2020, panwar_application_2020,bussolaAISlippingTiles2021}.   See \cite{varoquaux2022evaluating,thibeau2022clinicadl} for more details on data leakage.  Beyond procedures which are clearly wrong,  it can also detect lack of robustness to different parameters.  One would consider that the procedure is not statistically replicable if it leads to substantially different results under different train/test data splits,  different random seeds or small changes in hyperparameters.  Such an ML algorithm would display poor robustness and would be unlikely to be of future clinical use.  Note that, regarding the use of different train/test data splits, these would need to preserve a distribution of metadata  (for instance, age, sex, diagnosis\ldots) between train and test that is similar to that of the original study. Most classically, if the original study has stratified the splits, the statistical replication study would also need to stratify the splits. Using different distributions (for example not stratified) is also interesting but, in our view, falls within conceptual rather than statistical reproducibility. Furthermore,  it is very interesting to attempt replication on a different dataset with statistically equivalent characteristics: for instance another subsample which has not be used in the original study (but comes from the same larger dataset) or a different dataset but with similar characteristics (e.g.  same MRI scanners,  similar age,  similar disease stage\ldots).  Unsuccessful replication may be an indication of overfitting of the dataset of the original study through excessive experimentation with different architectures or hyper-parameters which ended up with a method that would work only on this very specific dataset.  This is referred to as the {\sl researcher degrees of freedom}~\cite{simmons2011false,gorgolewski2016practical}.  This concept extends beyond the field of ML.  It actually comes from experimental sciences where different statistical procedures are tried until a statistically significant result is found,  a bad practice known as {\sl p-hacking} \cite{head2015extent}.  It is important that researchers in our field have this problem in mind.  Experimental sciences have proposed {\sl preregistered} and {\sl registered} studies as a potential solution to ban such bad practices.  Preregistration means that the research plan is written down and made public before the study starts.  It can for example be published on the Open Science Foundation website\footnote{\url{https://osf.io}}.  This mechanism reduces the researcher degrees of freedom and is thus likely to lead to more robust results.  Registration goes one step further.  The research plan is submitted to a journal and peer-reviewed.  Thus (most of) the peer-review is done before the results are known.   It has the additional advantage of putting more focus on methodological soundness than on the ground-breaking nature of results (for instance negative results will be published).  More details about preregistration and registration can be found in \cite{henderson2022guide}.  Preregistration and registration are not yet widely used in ML for medical imaging.  Such practices would certainly not fit all studies because they leave no room for methodological creativity.  On the other hand, they should be very valuable to experimental studies aiming at validating ML methods.  We believe that, as a community,  we should try to adapt such procedures to our field.

\section{Conceptual reproducibility}
\label{sec:conceptual}
\paragraph{What is it?}
Conceptual reproducibility can be seen as the ultimate goal: the one which lead to the consolidation of scientific knowledge.  The general idea is to be able to validate the findings  under conceptually similar conditions\footnote{Again,  we use the term of \cite{mcdermott2019reproducibility} although with a slightly different meaning. }.  Conceptually similar means that the method,  the data and the experiments are compatible with the claims of the original study but they are not identical.  We will come back to the notion of claims of a study,  and their relationships to generalizability and limitations,  later in this section.

\paragraph{What does it require?}
Again,  we may distinguish between factors that make it possible to {\sl attempt} replication and those that will make it more likely to be {\sl successful}.

In theory,  nothing but the original paper should be strictly necessary.  Nevertheless,  this assumes that the original paper has adhered to the scientific gold-standard of providing all details necessary for replication: not only a description of the methods which makes reimplementation possible but a detailed description of the datasets and experimental procedure.  It is particularly worrisome that many medical imaging publications do not even report basic demographic statistics~\cite{varoquaux2022machine}.  \citet{gundersen2018state} argues that the replication should be independent of the implementation.  We agree in principle but believe that such requirement would considerably lower the number of conceptual replication attempts,  while more are needed to advance our field in a steadier manner.  In practice,  it is extremely useful to be able to access the code,  not only to save a lot of time but also to make sure that an unsuccessful replication is not due to a faulty reimplementation.  The same can be said for trained models.  Access to the original data can be useful to dissect the potential reasons for differences in results.  In summary,  none of the elements of exact reproducibility are required,  all of them are welcome.  

There are several characteristics of an original study that make it less likely for it to be replicated.  Low sample size not only means that it is less likely to find a true effect if it exists but also increases the odds that a positive finding is false~\cite{ioannidis2005most}.  This is not only true in ML but in experimental sciences in general.  Ideally,  the sample size should be justified by a previous power analysis  \cite{poldrack2017scanning}.  Causes for failure of statistical reproducibility also apply here.  In particular, too many researcher degrees of freedom increase the likelihood of having built a method that is overly specific to a dataset.  Another problem is that the datasets used in medical imaging ML papers are very often not representative of what would be found in the clinic~\cite{varoquaux2022machine}.  Indeed,  they often come from research datasets where the inclusion criteria are specific, the medical imaging protocols are harmonized and the data quality is controlled.  Thus,  it is necessary to have more studies including clinical routine data (e.g.  \cite{bottani2022automatic,perkuhn2018clinical}).  Finally, it is very important to have in mind that most scientific findings will not universally replicate but that the replication will only succeed under specific conditions.  This is why it is critical that scientific papers precisely define their claims and their limitations.  For instance,   a claim could be that a given algorithm can segment brain tumors with a Dice of $0.9\pm0.02$ when the MR images are acquired at 3 Tesla and have only minimal artifacts.  The same paper would mention as limitations that it is unclear how the algorithm would perform at 1.5 Tesla or with data of lower quality.  One can see that stating clear claims and limitations will allow defining the scope of conceptual replication studies.  Studies outside that scope would aim at studying generalizability beyond original claims.

\paragraph{Why is it useful?}
As mentioned above,  conceptual reproducibility is the ultimate goal,  the one which, through accumulation of evidence,  builds consensus about new scientific knowledge.  Its utility in general is thus obvious.  More specifically,  it provides different benefits.  In particular, in the field of ML for medical imaging, it allows studying the generalizability of a method.  It is thus a step towards its applicability to the clinic. To that aim,  the use of multiple datasets is of paramount importance.  This will not only allow ruling out that a method is overly specific to a given dataset.  It will allow defining which are the bounds within which the method applies. This includes the machine model, the acquisition parameters and the data quality.  It also includes factors which are unrelated to imaging such as population age, sex, geographic origin, disease severity and others.

\section{Measurement reproducibility}
\label{sec:measurement}
\paragraph{What is it?}
Measurement reproducibilty is the study of the variability of a specific measurement under different acquisition conditions. We are aware that,  at first sight, this concept does not fit
 ideally in our taxonomy (see Section~\ref{subsec:abouttypes} for a more detailed discussion).  Nevertheless, we chose to present it as a separate entity because this is a very common meaning of the word reproducibility in medical imaging~\footnote{Note that the word is used to evaluate reproducibility of automatic methods across different scans of the same subject but also when a human rater is involved (manual or semi-automated measurements), including intra-rater (measurement twice by the same rater) and inter-rater (two different raters) reproducibility from a single scan. } (e.g.   \cite{lukas2004sensitivity,borga2020reproducibility,chard2002reproducibility, de2010accuracy,lemieux1999fast,tudorascu2016reproducibility, yamashita2020radiomic,poldrack2019introduction}) and we thus believe that it deserves a special treatment.
Here,  we consider an algorithm that produces a measurement for each individual patient (for instance the volume of an anatomical structure computed by a segmentation method). A prototypical example of measurement reproducibility is the test-retest reproducibility: how much does the measure vary when applied to two different scans of the same patient? One can then introduce different variations: scans on the same day or not,  scans on the same or different machines,  systematic addition of noise or artifacts to the data\ldots. Finally, some authors call inter-method reproducibility the comparison of different software packages for the measurement of the same anatomical entity~\cite{palumbo2019evaluation}. We do not believe this falls within the topic of reproducibility but rather of methods' comparison~\footnote{It is of interest to compare which of them is the most accurate or robust, with respect to a ground truth. However, as mentioned above, we do not believe it falls within the topic of reproducibility.}.

\paragraph{What does it require?}
The code is necessary to make sure that variations do not depend on implementation and to ease the reproducibility study.  The trained models are also very welcome to facilitate the process.  It is then necessary to have access to test-retest data, meaning different acquisitions of the same patient.  As mentioned above: the more varied these different acquisitions,  the more extensive the study.  Ideally, one would want to have access to scans performed on the same day~\cite{lukas2004sensitivity, tudorascu2016reproducibility}, on different days~\cite{de2010accuracy, lemieux1999fast}, at different times during the day (for example before or after caffeine consumption, a factor which affects functional MRI measures~\cite{laurienti2002dietary}),  on different imaging devices~\cite{borga2020reproducibility}, with different acquisition parameters~\cite{yamashita2020radiomic}\ldots It is unlikely to obtain that many scans for the same patients.  A more feasible approach is to study these different factors of variations for different patients. Furthermore,  starting with a given image, it is possible to simulate different types of alterations and defects by adding them to the original image. This can be very useful because it allows generating very large numbers of images easily and to control for specific imaging characteristics (such as for example the level of noise or the strength of motion artifacts).  Such simulations can involve completely synthetic images called phantoms~\cite{collins1998design} which mimic real images. It can also be done through the addition of defects to real images~\cite{shawMRIKSpaceMotion,duffyRetrospectiveCorrectionMotion}.
Ideally,  measurement reproducibility should be performed in different populations of participants separately (for instance a child with autism spectrum disorder or a patient with Parkinson's disease are more likely to move during the acquisition and the image is thus more likely to be affected by motion artifacts).

\paragraph{Why is it useful?}
Measurement reproducibility is central for measurement sciences and medical imaging is one of those.  It is an extremely precious information to the user (for instance the radiologist).  Indeed,  it provides,  at the individual patient level,  and ideally for different categories of patients, the precision that they may expect from the measurement tool.  There is a wide tradition to perform such reproducibility studies in radiology journals. We believe that it would be very welcome that it becomes more commonplace in the ML for medical imaging community.

\section{Discussion}
\label{sec:discussion}

\subsection{About the different types of reproducibility}
\label{subsec:abouttypes}
We have presented different types of reproducibility. Our taxonomy is not original nor aims at being universal.  The boundaries between types are partly fuzzy.  For instance,  to which degree replication with a different but similar dataset should be considered {\sl statistical} or {\sl conceptual} reproducibility? We do not believe such questions to be of great importance.  Rather,  it is fruitful, following Gundersen and  Kjensmo~\cite{gundersen2018state} and Peng~\cite{peng2011reproducible}, to consider reproducibility as a spectrum.  In particular,  one can consider that the first three types provide increasing support for a finding: conceptual provides more support than statistical which in turns provides more support than exact.  The amount of components necessary to perform them is in the reverse order: exact requires more than statistical which requires more than conceptual.   Does it mean that only conceptual reproducibility matters? Absolutely not.  As we mentioned,  other types of reproducibility are necessary to dissect why a given replication has failed as well as to better specify the bounds within which a scientific claim is valid.  Last but not least,  exact reproducibility also helps build trust in science.

We must admit that measurement reproducibility does not fit very well in this landscape.  Moreover,  one could also argue that it is a type of conceptual reproducibility, which is partly true as it aims at studying the reproducibility when varying the input data.  We nevertheless believe it deserves a special treatment, for several reasons.  First,  here reproducibility is studied at the individual (i.e.  patient) level and not at the population level.  Also,  the emphasis is on the measurement rather than the finding.   Even if it has its role in the building of scientific knowledge, it has specific practical implications for the user.  Moreover, as mentioned above,  this is actually the most widely used meaning of reproducibility in medical imaging and it seemed important that the reader is acquainted with it.

\subsection{The many benefits of reproducibility}
\label{subsec:benefits}

 {\sl ``Der Weg ist das Ziel''} is a German saying which can be roughly translated as: ``the path is the goal''.  Indeed,  reproducibility allows researchers to discover many new places down the road before reaching the final destination. Even if this destination is never reached,  the benefits of the travel are of major importance.  Let us try to list some of them. 

There are many individual benefits for researchers and labs.  An important one is that aiming at reproducible research results in {\sl reusable} research artifacts.  How agreeable it is for a researcher to easily reuse an old code for a new project! How useful it is for a research lab to have data organized according to community standards making it easier to reuse and share! Moreover,  papers that come with shared data ~\cite{piwowar2007sharing,piwowar2013data,gorgolewski2016practical} or code attract~\cite{vandewalle2012code}, on average, more citations.  Thus aiming at reproducibility is also in the researchers' self interest.

There are also considerable benefits for the scientific community as a whole. As mentioned before,  reproducible research is often associated to open code, open data and available trained models.  This allows researchers not only to use them to perform replication studies but also to use these research artifacts for completely different purposes such as building new methods or conducting analysis on pooled datasets.  In the specific case of ML for medical imaging,  it also allows assessing independently the influence of preprocessing,  feature extraction and ML method.  This is particularly important when claims of superiority of a new ML method are made but the original paper uses overly specific preprocessing steps. 

Of course,  at the end of the path, the goal itself brings many benefits.  These have already largely described in the previous sections so we will just mention them briefly.  Conceptual replication studies are necessary for corroborating findings and thus building new scientific knowledge.  Statistical replication allows ensuring that results are not due to cherry picking.  Exact replication allows detecting errors and increases trust in science in general.

\subsection{One size does not fit all}
\label{subsec:onesize}
We hope the reader is now convinced of the benefits of aiming towards reproducible research.  Does it mean that reproducibility requirements should be the same for all studies? We strongly believe the opposite.  To take an extreme example,  requiring all studies to be exactly reproducible with minimal efforts (like with running a single command) would be an awful idea.  We believe,  on the contrary,  that reproducibility efforts should vary according to many factors including the type of study and the context in which it is performed.  One would probably not have the same level of requirement for a methodological paper and for an extensive medical application with strong claims about clinical applicability.  For the former, one may be satisfied with an experiment on a single or a few datasets.  For the later,  one would expect the study to include multiple datasets with varying characteristics and a comprehensive assessment of generalizability under different factors such as imaging devices and acquisition parameters.  Also,  there are some cases where sharing the code is not desired (e.g.  because an industrial application is foreseen) or where the code will not adhere to best development practices because it is just a prototype to test a new methodology.  Nevertheless, sharing weakly documented code is always better than no sharing at all. Similarly,  there are cases where data sharing is difficult or even impossible due to regulatory constraints. As mentioned above, reproducibility is a spectrum.  Where a given study should lie in this spectrum should depend on the type of study and the constraints the researchers face.

We thus advocate for a non-dogmatic approach to reproducibility.  Guidelines are extremely useful but they should not be carved in stone.  Also, we believe that the requirements should be assessed by the reviewers on a case by case basis.  Indeed,  what matters is that the reproducibility level matches the claims made in the paper.  Of course,  it is a good thing that journals and conferences provide requirements for reporting essential information.  It is helpful to researchers and makes the community progress towards better science.  Also,  some bad practices such as data leakage or p-hacking need to be banished.  But we believe that very high reproducibility requirements (e.g.  requiring that exact reproducibility is feasible) at the level of a given journal or conference would be counter-productive.  Finally,  we  like the idea of a badging system~\cite{heil2021reproducibility} which would tag papers according to their reproducibility level.  It remains to  be seen how such system should be implemented.

To conclude,  we firmly believe that it is essential for researchers and students in the field of ML for medical imaging to be trained to the concepts and practice of reproducibility.  It will be beneficial to them as well as to the community in general.  But this does not mean that researchers should aim at perfect reproducibility in all their studies.  Diversity in research approaches and practices is also a factor that drives science forward and which should be preserved.

% Acknowledgments section
\section*{Acknowledgments}
The authors are grateful to Gaël Varoquaux for pointing them towards useful references.
This work was supported by the French government under management of Agence
Nationale de la Recherche as part of the ``Investissements d'avenir''
program, reference ANR-19-P3IA-0001 (PRAIRIE 3IA Institute) and
ANR-10-IAIHU-06 (Agence Nationale de la Recherche-10-IA Institut
Hospitalo-Universitaire-6).
ETS acknowledges funding from the 4TU Precision Medicine program supported by High Tech for a Sustainable Future, a framework commissioned by the four Universities of Technology of the Netherlands.

% References section
\bibliographystyle{spbasicsort}
\bibliography{references}

\begin{thebibliography}{73}
\providecommand{\natexlab}[1]{#1}
\providecommand{\url}[1]{{#1}}
\providecommand{\urlprefix}{URL }
\expandafter\ifx\csname urlstyle\endcsname\relax
  \providecommand{\doi}[1]{DOI~\discretionary{}{}{}#1}\else
  \providecommand{\doi}{DOI~\discretionary{}{}{}\begingroup
  \urlstyle{rm}\Url}\fi
\providecommand{\eprint}[2][]{\url{#2}}

\bibitem[{Seab et~al(1988)Seab, Jagust, Wong, Roos, Reed, and
  Budinger}]{seab1988quantitative}
Seab J, Jagust W, Wong S, Roos M, Reed BR, Budinger T (1988) Quantitative {NMR}
  measurements of hippocampal atrophy in {A}lzheimer's disease. Magnetic
  Resonance in Medicine 8(2):200--208

\bibitem[{Lehericy et~al(1994)Lehericy, Baulac, Chiras, Pierot, Martin, Pillon,
  Deweer, Dubois, and Marsault}]{lehericy1994amygdalohippocampal}
Lehericy S, Baulac M, Chiras J, Pierot L, Martin N, Pillon B, Deweer B, Dubois
  B, Marsault C (1994) Amygdalohippocampal {MR} volume measurements in the
  early stages of {A}lzheimer disease. American Journal of Neuroradiology
  15(5):929--937

\bibitem[{Jack et~al(1997)Jack, Petersen, Xu, Waring, O'Brien, Tangalos, Smith,
  Ivnik, and Kokmen}]{jack1997medial}
Jack CR, Petersen RC, Xu YC, Waring SC, O'Brien PC, Tangalos EG, Smith GE,
  Ivnik RJ, Kokmen E (1997) Medial temporal atrophy on {MRI} in normal aging
  and very mild alzheimer's disease. Neurology 49(3):786--794

\bibitem[{Varoquaux and Colliot(2022)}]{varoquaux2022evaluating}
Varoquaux G, Colliot O (2022) Evaluating machine learning models and their
  diagnostic value. HAL preprint hal-03682454,
  \urlprefix\url{https://hal.archives-ouvertes.fr/hal-03682454/}

\bibitem[{Thibeau-Sutre et~al(2022)Thibeau-Sutre, Diaz, Hassanaly, Routier,
  Dormont, Colliot, and Burgos}]{thibeau2022clinicadl}
Thibeau-Sutre E, Diaz M, Hassanaly R, Routier A, Dormont D, Colliot O, Burgos N
  (2022) {ClinicaDL}: an open-source deep learning software for reproducible
  neuroimaging processing. Computer Methods and Programs in Biomedicine
  220:106818

\bibitem[{Simmons et~al(2011)Simmons, Nelson, and Simonsohn}]{simmons2011false}
Simmons JP, Nelson LD, Simonsohn U (2011) False-positive psychology:
  undisclosed flexibility in data collection and analysis allows presenting
  anything as significant. Psychological Sciences 22:1359–1366

\bibitem[{Baker(2016)}]{baker20161}
Baker M (2016) 1,500 scientists lift the lid on reproducibility. Nature
  533(7604)

\bibitem[{Gundersen(2020)}]{gundersen2020reproducibility}
Gundersen OE (2020) The reproducibility crisis is real. AI Magazine
  41(3):103--106

\bibitem[{Ioannidis(2005)}]{ioannidis2005most}
Ioannidis JP (2005) Why most published research findings are false. PLoS
  Medicine 2(8):e124

\bibitem[{Begley and Ioannidis(2015)}]{begley2015reproducibility}
Begley CG, Ioannidis JP (2015) Reproducibility in science: improving the
  standard for basic and preclinical research. Circulation research
  116(1):116--126

\bibitem[{Collaboration(2015)}]{open2015estimating}
Collaboration OS (2015) Estimating the reproducibility of psychological
  science. Science 349(6251):aac4716

\bibitem[{Begley(2013)}]{begley2013unappreciated}
Begley CG (2013) An unappreciated challenge to oncology drug discovery:
  pitfalls in preclinical research. American Society of Clinical Oncology
  Educational Book 33(1):466--468

\bibitem[{Sonnenburg et~al(2007)Sonnenburg, Braun, Ong, Bengio, Bottou, Holmes,
  LeCunn, Muller, Pereira, Rasmussen et~al}]{sonnenburg2007need}
Sonnenburg S, Braun ML, Ong CS, Bengio S, Bottou L, Holmes G, LeCunn Y, Muller
  KR, Pereira F, Rasmussen CE, et~al (2007) The need for open source software
  in machine learning. The Journal of Machine Learning Research

\bibitem[{Gundersen and Kjensmo(2018)}]{gundersen2018state}
Gundersen OE, Kjensmo S (2018) State of the art: Reproducibility in artificial
  intelligence. In: Proceedings of the AAAI Conference on Artificial
  Intelligence, vol~32

\bibitem[{Hutson(2018)}]{hutson2018artificial}
Hutson M (2018) Artificial intelligence faces reproducibility crisis. Science

\bibitem[{Haibe-Kains et~al(2020)Haibe-Kains, Adam, Hosny, Khodakarami,
  Waldron, Wang, McIntosh, Goldenberg, Kundaje, Greene
  et~al}]{haibe2020transparency}
Haibe-Kains B, Adam GA, Hosny A, Khodakarami F, Waldron L, Wang B, McIntosh C,
  Goldenberg A, Kundaje A, Greene CS, et~al (2020) Transparency and
  reproducibility in artificial intelligence. Nature 586(7829):E14--E16

\bibitem[{Stupple et~al(2019)Stupple, Singerman, and
  Celi}]{stupple2019reproducibility}
Stupple A, Singerman D, Celi LA (2019) The reproducibility crisis in the age of
  digital medicine. NPJ Digital Medicine 2(1):1--3

\bibitem[{McDermott et~al(2019)McDermott, Wang, Marinsek, Ranganath, Ghassemi,
  and Foschini}]{mcdermott2019reproducibility}
McDermott M, Wang S, Marinsek N, Ranganath R, Ghassemi M, Foschini L (2019)
  Reproducibility in machine learning for health. arXiv preprint
  arXiv:190701463

\bibitem[{Beam et~al(2020)Beam, Manrai, and Ghassemi}]{beam2020challenges}
Beam AL, Manrai AK, Ghassemi M (2020) Challenges to the reproducibility of
  machine learning models in health care. JAMA 323(4):305--306

\bibitem[{Gorgolewski and Poldrack(2016)}]{gorgolewski2016practical}
Gorgolewski KJ, Poldrack RA (2016) A practical guide for improving transparency
  and reproducibility in neuroimaging research. PLoS Biology 14(7):e1002506

\bibitem[{Nichols et~al(2017)Nichols, Das, Eickhoff, Evans, Glatard, Hanke,
  Kriegeskorte, Milham, Poldrack, Poline et~al}]{nichols2017best}
Nichols TE, Das S, Eickhoff SB, Evans AC, Glatard T, Hanke M, Kriegeskorte N,
  Milham MP, Poldrack RA, Poline JB, et~al (2017) Best practices in data
  analysis and sharing in neuroimaging using {MRI}. Nature Neuroscience
  20(3):299--303

\bibitem[{Poldrack et~al(2017)Poldrack, Baker, Durnez, Gorgolewski, Matthews,
  Munaf{\`o}, Nichols, Poline, Vul, and Yarkoni}]{poldrack2017scanning}
Poldrack RA, Baker CI, Durnez J, Gorgolewski KJ, Matthews PM, Munaf{\`o} MR,
  Nichols TE, Poline JB, Vul E, Yarkoni T (2017) Scanning the horizon: towards
  transparent and reproducible neuroimaging research. Nature Reviews
  Neuroscience 18(2):115--126

\bibitem[{Turkyilmaz-van~der Velden et~al(2020)Turkyilmaz-van~der Velden,
  Dintzner, and Teperek}]{turkyilmaz2020reproducibility}
Turkyilmaz-van~der Velden Y, Dintzner N, Teperek M (2020) Reproducibility
  starts from you today. Patterns 1(6):100099

\bibitem[{Heil et~al(2021)Heil, Hoffman, Markowetz, Lee, Greene, and
  Hicks}]{heil2021reproducibility}
Heil BJ, Hoffman MM, Markowetz F, Lee SI, Greene CS, Hicks SC (2021)
  Reproducibility standards for machine learning in the life sciences. Nature
  Methods 18(10):1132--1135

\bibitem[{Varoquaux(2018)}]{varoquaux2018cross}
Varoquaux G (2018) Cross-validation failure: Small sample sizes lead to large
  error bars. NeuroImage 180:68--77

\bibitem[{Button et~al(2013)Button, Ioannidis, Mokrysz, Nosek, Flint, Robinson,
  and Munaf{\`o}}]{button2013power}
Button KS, Ioannidis J, Mokrysz C, Nosek BA, Flint J, Robinson ES, Munaf{\`o}
  MR (2013) Power failure: why small sample size undermines the reliability of
  neuroscience. Nature Reviews Neuroscience 14(5):365--376

\bibitem[{Varoquaux and Cheplygina(2022)}]{varoquaux2022machine}
Varoquaux G, Cheplygina V (2022) Machine learning for medical imaging:
  methodological failures and recommendations for the future. NPJ Digital
  Medicine 5(1):1--8

\bibitem[{Bouthillier et~al(2019)Bouthillier, Laurent, and
  Vincent}]{bouthillier2019unreproducible}
Bouthillier X, Laurent C, Vincent P (2019) Unreproducible research is
  reproducible. In: International Conference on Machine Learning, PMLR, pp
  725--734

\bibitem[{Langer et~al(2018)Langer, Shih, Nagy, and
  Landman}]{langer2018collaborative}
Langer SG, Shih G, Nagy P, Landman BA (2018) Collaborative and reproducible
  research: goals, challenges, and strategies. Journal of Digital Imaging
  31(3):275--282

\bibitem[{Goodman et~al(2016)Goodman, Fanelli, and Ioannidis}]{goodman2016does}
Goodman SN, Fanelli D, Ioannidis JP (2016) What does research reproducibility
  mean? Science Translational Medicine 8(341):341ps12--341ps12

\bibitem[{Plesser(2018)}]{plesser2018reproducibility}
Plesser HE (2018) Reproducibility vs. replicability: a brief history of a
  confused terminology. Frontiers in Neuroinformatics 11:76

\bibitem[{McDermott et~al(2021)McDermott, Wang, Marinsek, Ranganath, Foschini,
  and Ghassemi}]{mcdermott2021reproducibility}
McDermott MB, Wang S, Marinsek N, Ranganath R, Foschini L, Ghassemi M (2021)
  Reproducibility in machine learning for health research: Still a ways to go.
  Science Translational Medicine 13(586):eabb1655

\bibitem[{Wilkinson et~al(2016)Wilkinson, Dumontier, Aalbersberg, Appleton,
  Axton, Baak, Blomberg, Boiten, da~Silva~Santos, Bourne
  et~al}]{wilkinson2016fair}
Wilkinson MD, Dumontier M, Aalbersberg IJ, Appleton G, Axton M, Baak A,
  Blomberg N, Boiten JW, da~Silva~Santos LB, Bourne PE, et~al (2016) The {FAIR}
  guiding principles for scientific data management and stewardship. Scientific
  Data 3(1):1--9

\bibitem[{Gabelica et~al(2022)Gabelica, Boj{\v{c}}i{\'c}, and
  Puljak}]{gabelica2022many}
Gabelica M, Boj{\v{c}}i{\'c} R, Puljak L (2022) Many researchers were not
  compliant with their published data sharing statement: mixed-methods study.
  Journal of Clinical Epidemiology

\bibitem[{Gorgolewski et~al(2016)Gorgolewski, Auer, Calhoun, Craddock, Das,
  Duff et~al}]{gorgolewski2016brain}
Gorgolewski KJ, Auer T, Calhoun VD, Craddock RC, Das S, Duff EP, et~al (2016)
  The brain imaging data structure, a format for organizing and describing
  outputs of neuroimaging experiments. Scientific Data 3(1):1--9

\bibitem[{Bourget et~al(2022)Bourget, Kamentsky, Ghosh, Mazzamuto, Lazari,
  Markiewicz, Oostenveld, Niso, Halchenko, Lipp et~al}]{bourget2022microscopy}
Bourget MH, Kamentsky L, Ghosh SS, Mazzamuto G, Lazari A, Markiewicz CJ,
  Oostenveld R, Niso G, Halchenko YO, Lipp I, et~al (2022) {Microscopy-BIDS}:
  An extension to the {Brain Imaging Data Structure} for microscopy data.
  Frontiers in Neuroscience 16

\bibitem[{Saborit-Torres et~al(2020)Saborit-Torres, Saenz-Gamboa, Montell,
  Salinas, G{\'o}mez, Stefan, Caparr{\'o}s, Garc{\'\i}a-Garc{\'\i}a, Domenech,
  Manj{\'o}n et~al}]{saborit2020medical}
Saborit-Torres J, Saenz-Gamboa J, Montell J, Salinas J, G{\'o}mez J, Stefan I,
  Caparr{\'o}s M, Garc{\'\i}a-Garc{\'\i}a F, Domenech J, Manj{\'o}n J, et~al
  (2020) Medical imaging data structure extended to multiple modalities and
  anatomical regions. arXiv preprint arXiv:201000434

\bibitem[{Cuingnet et~al(2011)Cuingnet, Gerardin, Tessieras, Auzias,
  Leh{\'e}ricy, Habert, Chupin, Benali, and Colliot}]{cuingnet2011automatic}
Cuingnet R, Gerardin E, Tessieras J, Auzias G, Leh{\'e}ricy S, Habert MO,
  Chupin M, Benali H, Colliot O (2011) Automatic classification of patients
  with {A}lzheimer's disease from structural {MRI}: a comparison of ten methods
  using the {ADNI} database. NeuroImage 56(2):766--781

\bibitem[{Samper-Gonz{\'a}lez et~al(2018)Samper-Gonz{\'a}lez, Burgos, Bottani,
  Fontanella, Lu, Marcoux, Routier, Guillon, Bacci, Wen
  et~al}]{samper2018reproducible}
Samper-Gonz{\'a}lez J, Burgos N, Bottani S, Fontanella S, Lu P, Marcoux A,
  Routier A, Guillon J, Bacci M, Wen J, et~al (2018) Reproducible evaluation of
  classification methods in {A}lzheimer's disease: Framework and application to
  {MRI and PET} data. NeuroImage 183:504--521

\bibitem[{Crane(2018)}]{crane2018questionable}
Crane M (2018) Questionable answers in question answering research:
  Reproducibility and variability of published results. Transactions of the
  Association for Computational Linguistics 6:241--252

\bibitem[{Carlini et~al(2021)Carlini, Tramer, Wallace, Jagielski, Herbert-Voss,
  Lee, Roberts, Brown, Song, Erlingsson et~al}]{carlini2021extracting}
Carlini N, Tramer F, Wallace E, Jagielski M, Herbert-Voss A, Lee K, Roberts A,
  Brown T, Song D, Erlingsson U, et~al (2021) Extracting training data from
  large language models. In: 30th USENIX Security Symposium (USENIX Security
  21), pp 2633--2650

\bibitem[{Abadi et~al(2016)Abadi, Chu, Goodfellow, McMahan, Mironov, Talwar,
  and Zhang}]{abadi2016deep}
Abadi M, Chu A, Goodfellow I, McMahan HB, Mironov I, Talwar K, Zhang L (2016)
  Deep learning with differential privacy. In: Proceedings of the 2016 ACM
  SIGSAC conference on computer and communications security, pp 308--318

\bibitem[{Gorgolewski et~al(2017)Gorgolewski, Alfaro-Almagro, Auer, Bellec,
  Capot{\u{a}}, Chakravarty, Churchill, Cohen, Craddock, Devenyi
  et~al}]{gorgolewski2017bids}
Gorgolewski KJ, Alfaro-Almagro F, Auer T, Bellec P, Capot{\u{a}} M, Chakravarty
  MM, Churchill NW, Cohen AL, Craddock RC, Devenyi GA, et~al (2017) {BIDS}
  apps: Improving ease of use, accessibility, and reproducibility of
  neuroimaging data analysis methods. PLoS Computational Biology 13(3):e1005209

\bibitem[{Routier et~al(2021)Routier, Burgos, D{\'\i}az, Bacci, Bottani,
  El-Rifai, Fontanella, Gori, Guillon, Guyot et~al}]{routier2021clinica}
Routier A, Burgos N, D{\'\i}az M, Bacci M, Bottani S, El-Rifai O, Fontanella S,
  Gori P, Guillon J, Guyot A, et~al (2021) Clinica: An open-source software
  platform for reproducible clinical neuroscience studies. Frontiers in
  Neuroinformatics 15

\bibitem[{McCormick et~al(2014)McCormick, Liu, Jomier, Marion, and
  Ibanez}]{mccormick2014itk}
McCormick M, Liu X, Jomier J, Marion C, Ibanez L (2014) {ITK}: enabling
  reproducible research and open science. Frontiers in Neuroinformatics 8:13

\bibitem[{Yoo et~al(2002)Yoo, Ackerman, Lorensen, Schroeder, Chalana, Aylward,
  Metaxas, and Whitaker}]{yoo2002engineering}
Yoo TS, Ackerman MJ, Lorensen WE, Schroeder W, Chalana V, Aylward S, Metaxas D,
  Whitaker R (2002) Engineering and algorithm design for an image processing
  {API}: a technical report on {ITK}-the insight toolkit. In: Medicine Meets
  Virtual Reality 02/10, IOS press, pp 586--592

\bibitem[{Drummond(2009)}]{drummond2009replicability}
Drummond C (2009) Replicability is not reproducibility: nor is it good science.
  In: Proceedings of the Evaluation Methods for Machine Learning Workshop at
  the 26th ICML, vol~1

\bibitem[{Bouthillier et~al(2021)Bouthillier, Delaunay, Bronzi, Trofimov,
  Nichyporuk, Szeto, Mohammadi~Sepahvand, Raff, Madan, Voleti
  et~al}]{bouthillier2021accounting}
Bouthillier X, Delaunay P, Bronzi M, Trofimov A, Nichyporuk B, Szeto J,
  Mohammadi~Sepahvand N, Raff E, Madan K, Voleti V, et~al (2021) Accounting for
  variance in machine learning benchmarks. Proceedings of Machine Learning and
  Systems 3:747--769

\bibitem[{Wen et~al(2020)Wen, Thibeau-Sutre, Diaz-Melo, Samper-González,
  Routier, Bottani, Dormont, Durrleman, Burgos, and
  Colliot}]{wen_convolutional_2020}
Wen J, Thibeau-Sutre E, Diaz-Melo M, Samper-González J, Routier A, Bottani S,
  Dormont D, Durrleman S, Burgos N, Colliot O (2020) Convolutional neural
  networks for classification of {Alzheimer}'s disease: {Overview} and
  reproducible evaluation. Medical Image Analysis 63:101694

\bibitem[{Samala et~al(2020)Samala, Chan, Hadjiiski, and
  Koneru}]{samala_hazards_2020}
Samala RK, Chan HP, Hadjiiski L, Koneru S (2020) Hazards of data leakage in
  machine learning: a study on classification of breast cancer using deep
  neural networks. In: Proc. SPIE Medical {Imaging} 2020: {Computer}-{Aided}
  {Diagnosis}, International Society for Optics and Photonics, vol 11314, p
  1131416

\bibitem[{Panwar et~al(2020)Panwar, Gupta, Siddiqui, Morales-Menendez, and
  Singh}]{panwar_application_2020}
Panwar H, Gupta PK, Siddiqui MK, Morales-Menendez R, Singh V (2020) Application
  of deep learning for fast detection of {COVID}-19 in {X}-{Rays} using
  {nCOVnet}. Chaos, Solitons \& Fractals 138:109944

\bibitem[{Bussola et~al(2021)Bussola, Marcolini, Maggio, Jurman, and
  Furlanello}]{bussolaAISlippingTiles2021}
Bussola N, Marcolini A, Maggio V, Jurman G, Furlanello C (2021) {AI} slipping
  on tiles: Data leakage in {Digital Pathology}. In: Del~Bimbo A, Cucchiara R,
  Sclaroff S, Farinella GM, Mei T, Bertini M, Escalante HJ, Vezzani R (eds)
  Pattern {{Recognition}}. {{ICPR International Workshops}} and {{Challenges}},
  {Springer International Publishing}, {Cham}, Lecture {{Notes}} in {{Computer
  Science}}, pp 167--182, \doi{10.1007/978-3-030-68763-2_13}

\bibitem[{Head et~al(2015)Head, Holman, Lanfear, Kahn, and
  Jennions}]{head2015extent}
Head ML, Holman L, Lanfear R, Kahn AT, Jennions MD (2015) The extent and
  consequences of p-hacking in science. PLoS biology 13(3):e1002106

\bibitem[{Henderson(2022)}]{henderson2022guide}
Henderson EL (2022) A guide to preregistration and registered reports. Preprint
  \urlprefix\url{https://osf.io/preprints/metaarxiv/x7aqr/download}

\bibitem[{Bottani et~al(2022)Bottani, Burgos, Maire, Wild, Str{\"o}er, Dormont,
  Colliot, Group et~al}]{bottani2022automatic}
Bottani S, Burgos N, Maire A, Wild A, Str{\"o}er S, Dormont D, Colliot O, Group
  AS, et~al (2022) Automatic quality control of brain {T1}-weighted magnetic
  resonance images for a clinical data warehouse. Medical Image Analysis
  75:102219

\bibitem[{Perkuhn et~al(2018)Perkuhn, Stavrinou, Thiele, Shakirin, Mohan,
  Garmpis, Kabbasch, and Borggrefe}]{perkuhn2018clinical}
Perkuhn M, Stavrinou P, Thiele F, Shakirin G, Mohan M, Garmpis D, Kabbasch C,
  Borggrefe J (2018) Clinical evaluation of a multiparametric deep learning
  model for glioblastoma segmentation using heterogeneous magnetic resonance
  imaging data from clinical routine. Investigative Radiology 53(11):647

\bibitem[{Lukas et~al(2004)Lukas, Hahn, Bellenberg, Rexilius, Schmid,
  Schimrigk, Przuntek, K{\"o}ster, and Peitgen}]{lukas2004sensitivity}
Lukas C, Hahn HK, Bellenberg B, Rexilius J, Schmid G, Schimrigk SK, Przuntek H,
  K{\"o}ster O, Peitgen HO (2004) Sensitivity and reproducibility of a new fast
  {3D} segmentation technique for clinical {MR}-based brain volumetry in
  multiple sclerosis. Neuroradiology 46(11):906--915

\bibitem[{Borga et~al(2020)Borga, Ahlgren, Romu, Widholm, Dahlqvist~Leinhard,
  and West}]{borga2020reproducibility}
Borga M, Ahlgren A, Romu T, Widholm P, Dahlqvist~Leinhard O, West J (2020)
  Reproducibility and repeatability of {MRI}-based body composition analysis.
  Magnetic Resonance in Medicine 84(6):3146--3156

\bibitem[{Chard et~al(2002)Chard, Parker, Griffin, Thompson, and
  Miller}]{chard2002reproducibility}
Chard DT, Parker GJ, Griffin CM, Thompson AJ, Miller DH (2002) The
  reproducibility and sensitivity of brain tissue volume measurements derived
  from an {SPM}-based segmentation methodology. Journal of Magnetic Resonance
  Imaging: An Official Journal of the International Society for Magnetic
  Resonance in Medicine 15(3):259--267

\bibitem[{de~Boer et~al(2010)de~Boer, Vrooman, Ikram, Vernooij, Breteler,
  van~der Lugt, and Niessen}]{de2010accuracy}
de~Boer R, Vrooman HA, Ikram MA, Vernooij MW, Breteler MM, van~der Lugt A,
  Niessen WJ (2010) Accuracy and reproducibility study of automatic {MRI} brain
  tissue segmentation methods. NeuroImage 51(3):1047--1056

\bibitem[{Lemieux et~al(1999)Lemieux, Hagemann, Krakow, and
  Woermann}]{lemieux1999fast}
Lemieux L, Hagemann G, Krakow K, Woermann FG (1999) Fast, accurate, and
  reproducible automatic segmentation of the brain in {T}1-weighted volume
  {MRI} data. Magnetic Resonance in Medicine: An Official Journal of the
  International Society for Magnetic Resonance in Medicine 42(1):127--135

\bibitem[{Tudorascu et~al(2016)Tudorascu, Karim, Maronge, Alhilali, Fakhran,
  Aizenstein, Muschelli, and Crainiceanu}]{tudorascu2016reproducibility}
Tudorascu DL, Karim HT, Maronge JM, Alhilali L, Fakhran S, Aizenstein HJ,
  Muschelli J, Crainiceanu CM (2016) Reproducibility and bias in healthy brain
  segmentation: comparison of two popular neuroimaging platforms. Frontiers in
  Neuroscience 10:503

\bibitem[{Yamashita et~al(2020)Yamashita, Perrin, Chakraborty, Chou, Horvat,
  Koszalka, Midya, Gonen, Allen, Jarnagin et~al}]{yamashita2020radiomic}
Yamashita R, Perrin T, Chakraborty J, Chou JF, Horvat N, Koszalka MA, Midya A,
  Gonen M, Allen P, Jarnagin WR, et~al (2020) Radiomic feature reproducibility
  in contrast-enhanced {CT} of the pancreas is affected by variabilities in
  scan parameters and manual segmentation. European Radiology 30(1):195--205

\bibitem[{Poldrack et~al(2019)Poldrack, Whitaker, and
  Kennedy}]{poldrack2019introduction}
Poldrack RA, Whitaker K, Kennedy DN (2019) Introduction to the special issue on
  reproducibility in neuroimaging. NeuroImage

\bibitem[{Palumbo et~al(2019)Palumbo, Bosco, Fantacci, Ferrari, Oliva, Spera,
  and Retico}]{palumbo2019evaluation}
Palumbo L, Bosco P, Fantacci M, Ferrari E, Oliva P, Spera G, Retico A (2019)
  Evaluation of the intra-and inter-method agreement of brain {MRI}
  segmentation software packages: A comparison between {SPM12 and FreeSurfer}
  v6. 0. Physica Medica 64:261--272

\bibitem[{Laurienti et~al(2002)Laurienti, Field, Burdette, Maldjian, Yen, and
  Moody}]{laurienti2002dietary}
Laurienti PJ, Field AS, Burdette JH, Maldjian JA, Yen YF, Moody DM (2002)
  Dietary caffeine consumption modulates {fMRI} measures. NeuroImage
  17(2):751--757

\bibitem[{Collins et~al(1998)Collins, Zijdenbos, Kollokian, Sled, Kabani,
  Holmes, and Evans}]{collins1998design}
Collins DL, Zijdenbos AP, Kollokian V, Sled JG, Kabani NJ, Holmes CJ, Evans AC
  (1998) Design and construction of a realistic digital brain phantom. IEEE
  transactions on medical imaging 17(3):463--468

\bibitem[{Shaw et~al(2018)Shaw, Sudre, Ourselin, and
  Cardoso}]{shawMRIKSpaceMotion}
Shaw R, Sudre C, Ourselin S, Cardoso MJ (2018) {MRI K-}space motion artefact
  augmentation: Model robustness and task-specific uncertainty. In: Medical
  Imaging with Deep Learning - MIDL 2018

\bibitem[{Duffy et~al(2018)Duffy, Zhang, Tang, and
  Zhao}]{duffyRetrospectiveCorrectionMotion}
Duffy BA, Zhang W, Tang H, Zhao L (2018) Retrospective correction of motion
  artifact affected structural {{MRI}} images using deep learning of simulated
  motion. In: Medical Imaging with Deep Learning - MIDL 2018

\bibitem[{Peng(2011)}]{peng2011reproducible}
Peng RD (2011) Reproducible research in computational science. Science
  334(6060):1226--1227

\bibitem[{Piwowar et~al(2007)Piwowar, Day, and Fridsma}]{piwowar2007sharing}
Piwowar HA, Day RS, Fridsma DB (2007) Sharing detailed research data is
  associated with increased citation rate. PLoS One 2(3):e308

\bibitem[{Piwowar and Vision(2013)}]{piwowar2013data}
Piwowar HA, Vision TJ (2013) Data reuse and the open data citation advantage.
  PeerJ 1:e175

\bibitem[{Vandewalle(2012)}]{vandewalle2012code}
Vandewalle P (2012) Code sharing is associated with research impact in image
  processing. Computing in Science \& Engineering 14(4):42--47

\end{thebibliography}

\end{document}